\documentclass[10pt,twocolumn,letterpaper]{article}

\usepackage[pagenumbers]{cvpr} 
\usepackage{amsmath}
\newcommand{\Exp}{\operatorname{Exp}}
\definecolor{cvprblue}{rgb}{0.21,0.49,0.74}
\usepackage[pagebackref,breaklinks,colorlinks,allcolors=cvprblue]{hyperref}
\usepackage{multirow}

\def\confName{CVPR}
\def\confYear{2026}

\title{Dual-Agent Reinforcement Learning for Adaptive and Cost-Aware Visual–Inertial Odometry}

\author{%
 \textbf{Feiyang Pan}$^1$$^\dagger$,
 \textbf{Shenghe Zheng}$^{2}$$^\dagger$,
 \textbf{Chunyan Yin}$^{1*}$,
 \textbf{Guangbin Dou}$^{1}$
 \thanks{Corresponding Author(gdou@seu.edu.cn).~~~$^\dagger$Equal Contribution.}\\ 
  $^1$ Southeast University \qquad $^2$ Harbin Institute of Technology \\
  {\tt\small 230238437@seu.edu.cn}
}

\begin{document}
\maketitle
\begin{abstract}

Visual-Inertial Odometry (VIO) is a critical component for robust ego-motion estimation, enabling foundational capabilities such as autonomous navigation in robotics and real-time 6-DoF tracking for augmented reality.
%
Existing methods face a well-known trade-off: filter-based approaches are efficient but prone to drift, while optimization-based methods, though accurate, rely on computationally prohibitive Visual-Inertial Bundle Adjustment (VIBA) that is difficult to run on resource-constrained platforms.
Rather than removing VIBA altogether, we aim to reduce how often and how heavily it must be invoked. 
To this end, we cast two key design choices in modern VIO, when to run the visual frontend and how strongly to trust its output, as sequential decision problems, and solve them with lightweight reinforcement learning (RL) agents. Our framework introduces a lightweight, dual-pronged RL policy that serves as our core contribution: (1) a Select Agent intelligently gates the entire VO pipeline based only on high-frequency IMU data; and (2) a composite Fusion Agent that first estimates a robust velocity state via a supervised network , before an RL policy adaptively fuses the full (p, v, q) state.
%
Experiments on the EuRoC MAV and TUM-VI datasets show that, in our unified evaluation, the proposed method achieves a more favorable accuracy–efficiency–memory trade-off than prior GPU-based VO/VIO systems: it attains the best average ATE while running up to $1.77\times$ faster and using less GPU memory. 
Compared to classical optimization-based VIO systems, our approach maintains competitive trajectory accuracy while substantially reducing computational load.
\end{abstract}    

\section{Introduction}
\label{sec:intro}

Accurate, robust, real-time ego-motion estimation is fundamental for autonomous systems, underpinning navigation for mobile robots, UAVs, and self-driving cars in GPS-denied environments, as well as real-time 6-DoF tracking for AR\cite{10492667,PANIGRAHI20226019,2023Augmented}. Visual-Inertial Odometry (VIO) has become a dominant paradigm, fusing visual and inertial data to deliver high-frequency, metrically scaled pose estimates. A typical VIO pipeline processes high-rate IMU measurements (via pre-integration) and low-rate visual frames (for feature tracking), which are then reconciled by a backend estimator into a single coherent state.
The backend estimator for this reconciliation has historically been dominated by two paradigms. Filter-based approaches\cite{1961New,2013High,2015Robust} offer high computational efficiency for real-time deployment, but their accuracy is sub-optimal due to linearization errors and noise accumulation. 
Conversely, optimization-based methods achieve superior 
accuracy by performing non-linear Visual-Inertial Bundle Adjustment (VIBA)\cite{qin2018vins,qin2019general,campos2021orb,von2022dm}. However, the substantial computational burden of this backend VIBA optimization makes it challenging to deploy on resource-constrained edge devices .
Recent advancements in deep learning (DL) have attempted to 
alleviate this trade-off. However, end-to-end DL models\cite{almalioglu2022selfvio,shamwell2019unsupervised,clark2017vinet}, 
while novel, still exhibit limitations in precision and generalization 
compared to mature optimization frameworks. 
Hybrid methods have also emerged, integrating learned components 
for robustness\cite{pan2024adaptive,fu2024islam}. While promising, 
these hybrid approaches are still ultimately constrained by the exact 
same computational VIBA bottleneck they build upon . 
As a result, the accuracy–efficiency trade-off remains a practical challenge for real-time VIO on resource-constrained platforms.
\begin{figure}[t]
  \centering
  
   \includegraphics[width=1\linewidth]{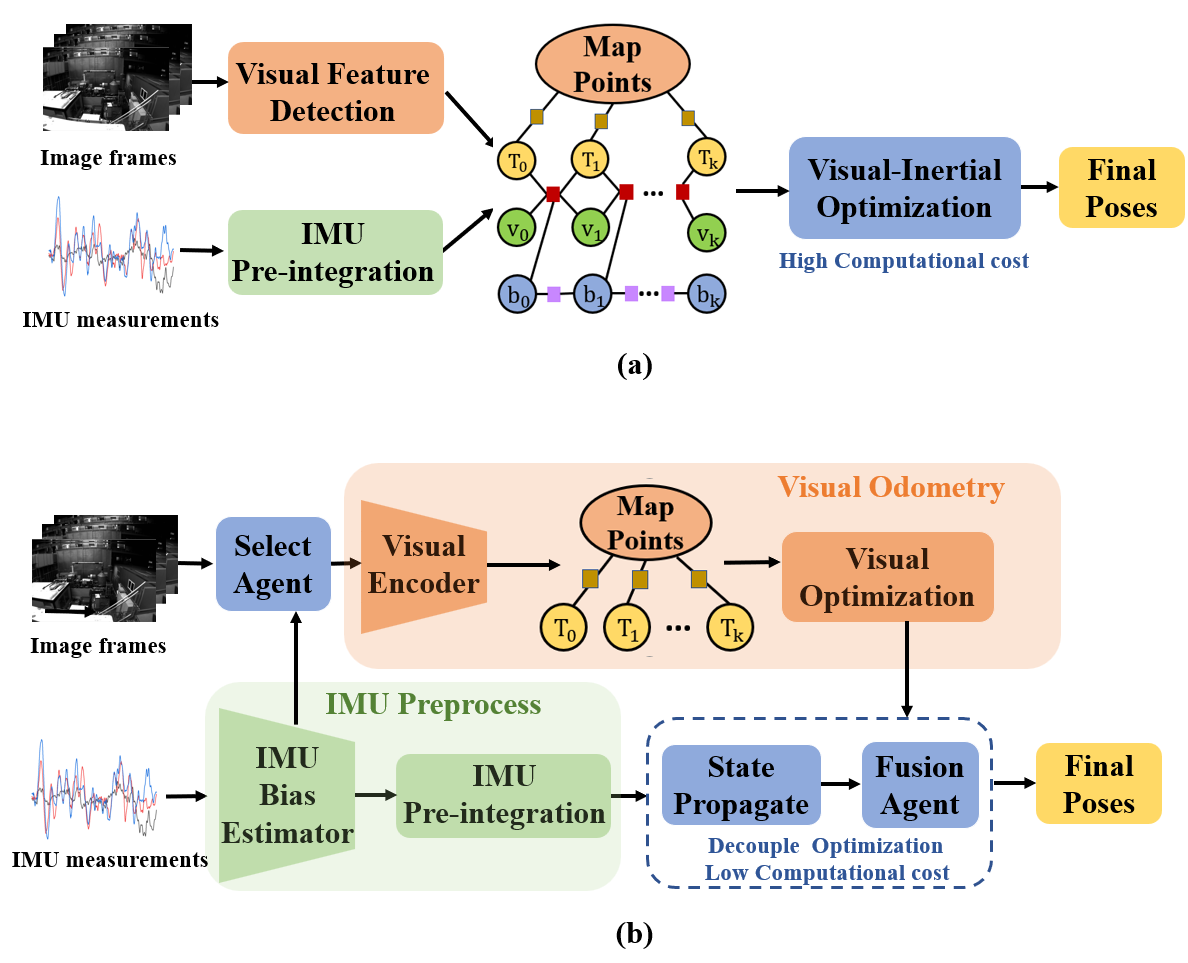}

   \caption{The accuracy-efficiency trade-off in VIO. 
(a) The traditional tightly-coupled VIO framework, which relies on a monolithic and computationally expensive Visual-Inertial Bundle Adjustment (VIBA) block.
(b) Our proposed decoupled RL-based framework. We mitigate the VIBA bottleneck by introducing two intelligent agents.}
   \label{fig:onecol}
\end{figure}

Rather than attempting to remove VIBA altogether, our goal is to reduce how often and how heavily it must be invoked. 
To this end, we focus on two key design choices in modern VIO: 
(i) when to run the visual frontend, and 
(ii) how strongly to trust its output relative to IMU-based propagation. 
We view these choices not as fixed heuristics or implicit byproducts of an optimizer, but as explicit sequential decisions. 
Casting them as a decision process naturally leads to a reinforcement learning (RL) formulation: a policy can learn, from data, to trigger costly VO updates only when needed and to adapt fusion weights according to motion intensity and visual reliability.
As depicted in Figure~\ref{fig:onecol}(b), we introduce 
a novel, dual-pronged RL policy . This policy 
features a Select Agent 
that learns an intelligent scheduling policy; crucially, it 
makes an a prior decision based purely on 
high-frequency IMU data, allowing it to skip 
the entire costly VO pipeline on redundant frames. 
This is complemented by a composite Fusion Agent 
that first employs a supervised network 
(MLP1) to estimate a robust 
velocity state, after which its core RL policy (MLP2) 
learns to adaptively fuse the full (p, v, q) state. 
This dual-RL approach mitigates the monolithic, costly VIBA 
with a flexible, low-overhead, and intelligent decision-making 
process, thereby mitigating the primary computational bottleneck. In summary, the main contributions of this work can be summarized as follows: 

\noindent{\normalsize$\bullet$}\;
We cast two key VIO design problems, VO scheduling and visual–inertial fusion, as sequential decision tasks and solve them with lightweight RL agents. 
Instead of replacing the VO backend, this formulation learns a long-horizon, cost-aware policy that reduces the reliance on heavy VIBA.

\noindent{\normalsize$\bullet$}\;
We design a dual-agent RL architecture consisting of (i) an IMU-only Select Agent that reduces computational load by deciding when to run the entire VO pipeline, and (ii) a composite Fusion Agent that combines a supervised velocity estimator with an RL-based full-state fusion policy, improving the accuracy–efficiency trade-off.

\noindent{\normalsize$\bullet$}\;
We validate our framework on the EuRoC MAV and TUM-VI datasets, where it substantially reduces computational overhead while maintaining strong accuracy. 
In our unified evaluation, it achieves a better accuracy–throughput–memory trade-off than prior GPU-based VO/VIO systems and remains competitive with classical optimization-based VIO methods.
\section{Related Works}
\label{sec:rw}
\subsection{Classical VIO Paradigms}



Classical VIO approaches are largely divided into two paradigms. Filter-based methods, such as the seminal MSCKF\cite{2013High} and ROVIO\cite{2015Robust}, are highly computationally efficient, but their accuracy is often sub-optimal due to linearization errors and noise accumulation. Conversely, optimization-based methods\cite{qin2019general,von2022dm,von2018direct}, popularized by VINS-Mono\cite{qin2018vins} and advanced by systems like ORB-SLAM3\cite{campos2021orb}, achieve state-of-the-art accuracy by performing non-linear Visual-Inertial Bundle Adjustment (VIBA). This superior precision, however, comes at the cost of a significant computational burden, which remains a primary bottleneck for deployment on resource-constrained edge devices. Our work aims to resolve this fundamental trade-off.

\subsection{Deep Learning in VO/VIO}

The advent of deep learning (DL) has inspired two new VIO paradigms. End-to-end systems, such as VINet\cite{clark2017vinet}, BotVIO\cite{11024235} and SelfVIO\cite{almalioglu2022selfvio}, attempt to regress ego-motion directly from raw sensor inputs, but have generally struggled to match the precision and generalization of classical methods\cite{Han2019DeepVIOSD}. Consequently, a hybrid paradigm has emerged, integrating DL modules into traditional pipelines\cite{zhou2024dba,bloesch2018codeslam}. Works like iSLAM\cite{fu2024islam} and DPVO\cite{teed2023deep} use learned components to enhance feature matching or backend optimization\cite{teed2021droid,pan2024adaptive}. 
Several methods, such as RNIN-VIO~\cite{9583805} and CoVIO\cite{Vodisch_2023_CVPR} learn inertial or fusion components on large-scale driving datasets and and evaluate primarily on KITTI outdoor benchmarks\cite{Menze2015CVPR}. More closely related to our setting, adaptive visual-modality selection approaches~\cite{yang2022efficient,pan2024adaptive} dynamically re-weight visual and inertial cues inside deep VIO pipelines to improve robustness. However, these methods operate at the feature or latent representation level and still run the visual encoder on every frame, targeting accuracy and adaptivity rather than explicitly optimizing compute. In contrast, we target fully 6-DoF MAV and handheld scenarios with aggressive motion, and introduce a cost-aware decision layer on top of a VO backend that (i) makes IMU-only scheduling decisions before any visual computation and (ii) learns fusion weights that jointly account for motion, VO reliability, and computational budget.

\subsection{Reinforcement Learning in VO/VIO}
Reinforcement Learning (RL) has been successfully applied to high-level robotics tasks like autonomous exploration\cite{cimurs2021goal,cao2024deep,chaplot2020learning}, and more recently, to low-level control within odometry pipelines. A highly relevant work by Messikommer et al. reframes Visual Odometry (VO) as a sequential decision-making task, using an RL agent within the VO pipeline to replace internal heuristics like keyframe selection\cite{messikommer2024reinforcement}.
Our work differs in two key ways. 
First, we use RL not to tune internal VO heuristics given already-computed visual features, but as a high-level, IMU-only scheduler that decides whether to run the entire VO pipeline at a given time step. 
This allows the policy to skip the costly visual optimization component altogether when IMU integration is sufficient, yielding much larger computational savings than decisions that are conditioned on image features which have already been extracted. 
Second, we go beyond pure VO and address the full visual–inertial fusion problem: a second RL agent learns to adaptively fuse high-frequency IMU predictions with sparse VO updates using uncertainty-aware state representations. 

\section{Method}

\subsection{Motivation}
\label{sec:motivation}



Our design is guided by three practical goals for improving the accuracy–efficiency trade-off in VIO.

First, we aim to reduce reliance on tightly coupled visual–inertial bundle adjustment (VIBA). 
While accurate, VIBA is expensive for resource-constrained platforms, whereas loosely coupled schemes with fixed-gain filters~\cite{6906584,9714038} are efficient but cannot adapt well to changing motion and sensor conditions. 
Instead of using a fixed filter, we introduce a learned fusion policy and cast fusion as a sequential decision problem, enabling RL to optimize a long-horizon, cost-aware policy.

Second, we seek to avoid running the costly VO pipeline on redundant or uninformative frames. 
Existing keyframe selection strategies typically require image processing before deciding whether a frame is useful. 
In contrast, our Select Agent makes a priori scheduling decisions using only high-frequency IMU signals, allowing the system to skip entire VO passes and save computation.

Third, our dual-agent strategy depends on accurate IMU state predictions. 
Because IMU integration is highly sensitive to sensor bias, we introduce a dedicated IMU Bias Estimator network that provides pre-compensated inertial measurements as a reliable foundation for both the Select and Fusion agents.

\begin{figure*}[t]
  \centering
  
   \includegraphics[width=1\linewidth]{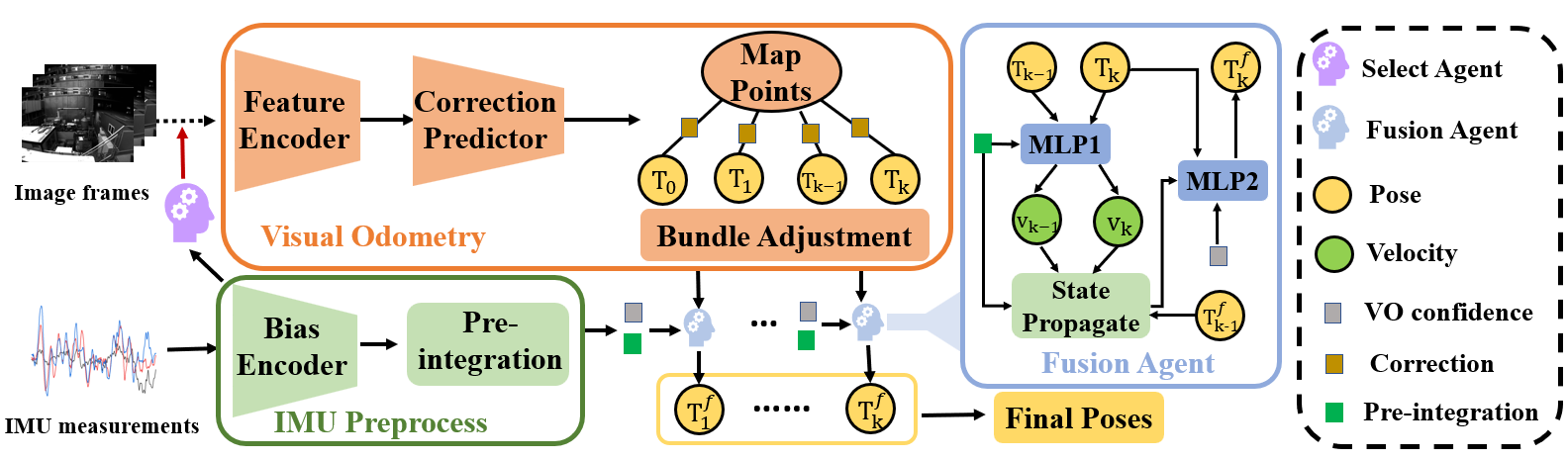}

   \caption{Overview of our proposed VIO pipeline. The system is composed of four decoupled modules: (1) \textbf{IMU Preprocess}, (2) \textbf{Select Agent}, (3) \textbf{Visual Odometry}, and (4) \textbf{Fusion Agent}. This framework leverages Reinforcement Learning to intelligently schedule and fuse sensor data, offering a highly computationally efficient alternative to traditional, tightly-coupled Visual-Inertial Bundle Adjustment.}
   \label{fig:method_overview}
\end{figure*}

\subsection{System Overview}
\label{sec:overview}

Our overall framework, illustrated in Fig.~\ref{fig:method_overview}, consists of four interacting modules. 
The \textbf{IMU Preprocess} module (Sec.~\ref{sec:bias_estimator}) uses a Bias Encoder followed by pre-integration to produce inter-frame inertial states $(\Delta \mathbf{p}, \Delta \mathbf{q}, \Delta \mathbf{v}, \Delta t)$. 
These states are fed to the \textbf{Select Agent} (Sec.~\ref{sec:rl_scheduling}), an RL policy that decides whether to activate the high-cost \textbf{Visual Odometry} (VO) module (Sec.~\ref{sec:vo_module}). 
When invoked, the DPVO-inspired VO backend outputs a sparse, high-accuracy but non-metric pose $T_k$, which is converted to a metric observation using a scale $s$ recovered during initialization (Sec.~\ref{sec:rl_fusion}). 
The \textbf{Fusion Agent} (Sec.~\ref{sec:rl_fusion}) then takes the scaled pose, VO confidence, the IMU pre-integration result, and the previous fused pose $T_{k-1}^f$ to produce the final pose $T_k^f$. 
This fused pose is fed back to the state propagation module as the starting point for the next cycle, closing the loop.
\subsection{IMU Bias Estimator}
\label{sec:bias_estimator}
A high-fidelity IMU prediction is foundational to our framework. We train a bias estimator rather than a signal denoiser, as an estimator is more adept at modeling the dominant, slowly-varying bias patterns and is less prone to overfitting\cite{10373097,10.5555/3504035.3504827,pan2025moe}. Our architecture consists of two lightweight encoder networks, $f_{bias}^g$ (for gyro) and $f_{bias}^a$ (for accelerometer). Each network ingests a window of raw, noisy sensor data along with the sensor's calibrated intrinsic parameters (noise density and bias stability). The gyro estimator $f_{bias}^g$ inputs the raw angular velocity sequence $\tilde{\Omega} = \{\tilde{\omega}_k\}$ and its noise profile $\mathbf{n}_g$, outputting a single 3-axis gyro bias estimate $\hat{\mathbf{b}}_g = f_{bias}^g(\tilde{\Omega}, \mathbf{n}_g)$. Similarly, $f_{bias}^a$ inputs the raw acceleration sequence $\tilde{\mathbf{A}} = \{\tilde{a}_k\}$ and its noise profile $\mathbf{n}_a$, outputting $\hat{\mathbf{b}}_a = f_{bias}^a(\tilde{\mathbf{A}}, \mathbf{n}_a)$. Training details in Appendix~A.




\noindent\textbf{Pre-integration\cite{article11}.}
At inference, the pre-trained $f_{bias}^g$ and $f_{bias}^a$ networks are used to correct the raw IMU measurements between two consecutive camera frames ($t_k, t_{k+1}$). These corrected measurements, $\hat{\omega}(k)$ and $\hat{a}(k)$, are then fed into a standard IMU pre-integration process . This module's final output, which is passed to the RL agents, consists of the relative motion $\Delta\mathbf{p}, \Delta\mathbf{q}, \Delta\mathbf{v}$ and the estimated bias $\hat{\mathbf{b}}$ for that interval. The detailed IMU pre-integration equations are provided in Appendix B.

\subsection{RL-based Computational Scheduling}
\label{sec:rl_scheduling}

We treat VO scheduling as a sequential decision problem under uncertainty: activating the VO pipeline improves accuracy but increases computation, while skipping it saves resources at the risk of drift. 
We therefore formulate scheduling as a Markov Decision Process (MDP)~\cite{1957A} and learn a long-horizon, cost-aware observation policy via reinforcement learning. During training, the Select Agent interacts with a log-driven but closed-loop simulator. 

\noindent{\normalsize$\bullet$}\;\textbf{State Space ($\mathcal{S}_{sel}$):} 
The Select Agent observes a compact IMU-only state
$s_t^{\text{sel}} = \{\Delta \mathbf{p}_t, \Delta \mathbf{q}_t, \Delta \mathbf{v}_t, \Delta t_t^{\text{vo}}\}$,
which summarizes pose/velocity drift and the elapsed time since the last VO update. 
This allows scheduling decisions to be made purely from inertial signals, without computing any visual features.

\noindent{\normalsize$\bullet$}\;\textbf{Action Space ($\mathcal{A}_{sel}$):} 
The action $a_t^{\text{sel}} \in \{0 \text{ (Skip VO)}, 1 \text{ (Run VO)}\}$ determines whether to invoke the VO pipeline at time $t$. 
This binary design enables complete omission of the visual encoder on skipped frames, while still allowing adaptive activation when drift risk increases.

\noindent{\normalsize$\bullet$}\;\textbf{Reward Function ($\mathcal{R}_{\text{sel}}$):}
We combine a terminal, ATE-based objective with a dense per-step shaping term. 
At the end of each episode, we assign
\begin{equation} \label{eq:reward_sel}
    R_{\text{episode}} = \frac{A}{\text{ATE} + \epsilon} - B \cdot N_{f},
\end{equation}
where $N_f$ is the number of VO calls in the episode, encouraging low trajectory error with few VO invocations.
During rollouts, each step $t$ additionally receives a small penalty that depends on the instantaneous pose error, which stabilizes learning. 
We train the agent with PPO~\cite{schulman2017proximalpolicyoptimizationalgorithms}; shaping coefficients, clipping ranges, and other hyperparameters are detailed in Appendix~A.

\noindent{\normalsize$\bullet$}\;\textbf{Policy Network:} 
The Select Agent policy $\pi_{\text{sel}}(a_t^{\text{sel}} \mid s_t^{\text{sel}})$ is parameterized by a small multi-layer perceptron (MLP).

\subsection{Visual Odometry Module}
\label{sec:vo_module}

This module (Fig.~\ref{fig:method_overview}) is activated only when the Select Agent(Sec.~\ref{sec:rl_scheduling}) outputs $a_t^{sel} = 1$. Its architecture is heavily inspired by the patch-based, recurrent optimization principles of \textbf{Deep Patch Visual Odometry (DPVO)}\cite{teed2023deep}. The process, following our system diagram, is as follows:

\begin{enumerate}
    \item \textbf{Feature Encoder:} A CNN processes the input image $I_t$ to extract dense feature maps , which are used for both context and matching.
    
    \item \textbf{Correction Predictor:} This module takes the feature maps from the Feature Encoder and the current state (poses $\mathbf{T}$, patch depths $\mathbf{P}$) as input. It first reprojects patch centers $\hat{P}_{kt}^{\prime} = \omega_{it}(\mathbf{T}, \mathbf{P}_k)$ into the current frame. It then computes visual similarity (correlation)  by comparing the patch's original features against the matching features $\mathcal{F}_t^m$ sampled at $\hat{P}_{kt}^{\prime}$.This correlation signal, along with the context features $\mathcal{F}_t^c$, is fed into its recurrent network, which outputs the 2D correction target $\delta_{kt} \in \mathbb{R}^2$ and a confidence weight $\Sigma_{kt} \in \mathbb{R}^2$ .
    
    \item \textbf{Map Points:} The Map Points (patches $\mathbf{P}$) and the Poses ($T_k$) form the set of variables to be optimized. The correction targets ($\delta, \Sigma$) define the factors in a factor graph, representing the optimization objective.
    
    \item \textbf{Bundle Adjustment:} This differentiable layer minimizes the objective function defined by these factors, solving for the optimal pose and depth updates ($\Delta \mathbf{T}, \Delta \mathbf{P}$):
    \begin{equation} \label{eq:vo_ba}
        \Delta \mathbf{T}, \Delta \mathbf{P} = \arg \min \sum_{(k,t)} \left\| \hat{\omega}_{it}(\mathbf{T}, \mathbf{P}_k) - [\hat{P}_{kt}^{\prime} + \delta_{kt}] \right\|_{\Sigma_{kt}}^2
    \end{equation}
\end{enumerate}

The final optimized pose $\mathbf{z}^{V}_t = \mathbf{T}_t^{init} \cdot \Delta \mathbf{T}$ serves as the high-accuracy observation for our fusion policy.
\subsection{RL-based Adaptive Fusion}
\label{sec:rl_fusion}
\begin{figure}[t]
  \centering
  
   \includegraphics[width=1\linewidth]{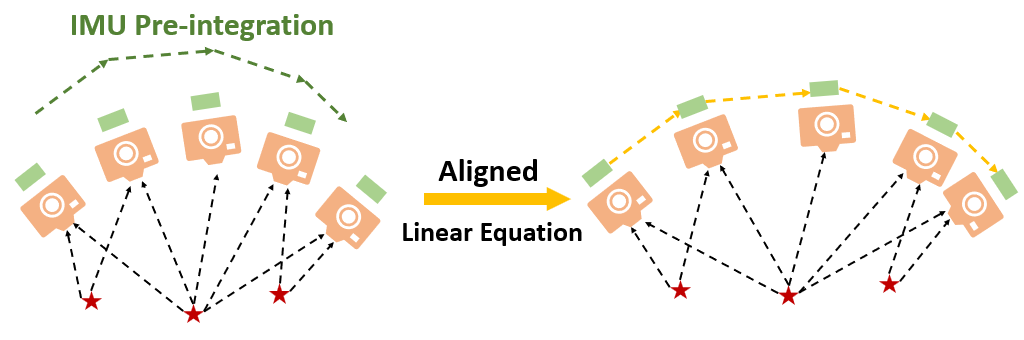}

   \caption{Visual summary of   initialization. 
   }
   \label{fig:initial}
\end{figure}

We fuse high-rate IMU state propagation with sparse, high-accuracy VO. To resolve the monocular scale ambiguity, a single global scale is estimated at start-up before fusion proceeds.

\noindent\textbf{1. System Initialization (R, T, S).}
We first align the world frame $W$ with the initial IMU body frame $b_0$, setting $\mathbf{p}_0^w=\mathbf{0}$ and aligning the $z$-axis with the measured gravity $\mathbf{g}$. 
This fixes the initial pose $(\mathbf{R}_0^w,\mathbf{T}_0^w)$.

To estimate the metric scale $s$ and the initial body-frame velocities $\mathbf{v}_{b_k}^w$, we adapt the linear initialization of VINS-Mono. 
Over a sliding window of $n$ keyframes (Fig.~\ref{fig:initial}), we collect IMU pre-integrations $(\boldsymbol{\alpha},\boldsymbol{\beta})$ and VO-derived relative translations $\mathbf{p}_{vo}^{(k,k+1)}$.

The metric IMU propagation provides
\begin{equation}
\label{eq:deltav}
\mathbf{v}_{b_{k+1}}^{w} 
= \mathbf{v}_{b_k}^{w} - \mathbf{g}^{w}\Delta t_k + \mathbf{R}_{b_k}^{w\top}\boldsymbol{\beta}_{k,k+1},
\end{equation}
\begin{equation}
\label{eq:deltap}
\mathbf{p}_{b_{k+1}}^{w} 
= \mathbf{p}_{b_k}^{w} + \mathbf{v}_{b_k}^{w}\Delta t_k - \tfrac{1}{2}\mathbf{g}^{w}\Delta t_k^2 
  + \mathbf{R}_{b_k}^{w\top}\boldsymbol{\alpha}_{k,k+1}.
\end{equation}

\noindent\emph{Camera pose from IMU via extrinsics.}
We obtain the camera orientation in the world, $\mathbf{R}_{c_k}^{w}$, from the IMU-derived body orientation $\mathbf{R}_{b_k}^{w}$ using the fixed camera–IMU extrinsics $(\mathbf{R}_{bc},\mathbf{t}_{bc})$ (body w.r.t.\ camera):
\[
\mathbf{R}_{c_k}^{w}=\mathbf{R}_{b_k}^{w}\mathbf{R}_{bc}^{\top}, 
\qquad 
\mathbf{p}_{c_k}^{w}=\mathbf{p}_{b_k}^{w}+\mathbf{R}_{b_k}^{w}\mathbf{t}_{bc}.
\]
The extrinsics are obtained offline (e.g., Kalibr) and kept fixed during all experiments.

\noindent\emph{Scale–translation coupling.}
Relating IMU-propagated positions to scaled VO translations gives the body–to–body displacement:
\begin{equation}
\mathbf{p}_{b_{k+1}}^{w}-\mathbf{p}_{b_k}^{w}
= \mathbf{R}_{c_k}^{w}\big(s\,\mathbf{p}_{vo}^{(k,k+1)}\big) 
  + \big(\mathbf{R}_{c_{k+1}}^{w}-\mathbf{R}_{c_k}^{w}\big)\mathbf{t}_{bc}.
\end{equation}
Substituting \eqref{eq:deltap} yields, for each pair $(k,k+1)$, the linear constraint
\begin{equation}
\label{eq:scale_linear}
\begin{aligned}
\mathbf{R}_{c_k}^{w}\mathbf{p}_{vo}^{(k,k+1)}\, s 
- \Delta t_k\,\mathbf{I}\,\mathbf{v}_{b_k}^{w}
= \;& \mathbf{R}_{b_k}^{w\top}\boldsymbol{\alpha}_{k,k+1}
     - \tfrac{1}{2}\mathbf{g}^{w}\Delta t_k^2 \\
&\; - \big(\mathbf{R}_{c_{k+1}}^{w}-\mathbf{R}_{c_k}^{w}\big)\mathbf{t}_{bc}.
\end{aligned}
\end{equation}

Stacking \eqref{eq:deltav}–\eqref{eq:scale_linear} over $n$ keyframes forms an over-determined linear system 
$\mathbf{H}\mathcal{X}=\mathbf{b}$ with $\mathcal{X}=[\mathbf{v}_{b_0}^w,\dots,\mathbf{v}_{b_n}^w,s]^\top$.
We solve it in least squares to obtain a single global metric scale $s$. 
Subsequently, all VO poses are scaled by $s$ and aligned with the IMU-derived orientation before entering the fusion pipeline.

\noindent\textbf{2. RL-based Adaptive Fusion Policy}
\label{sec:fusion_agent_rl}
The Fusion Agent addresses the challenge of dynamically balancing high-rate inertial propagation and low-rate visual updates. Unlike static fusion schemes, it learns a long-term policy that adapts the sensor weighting according to the evolving uncertainty and motion dynamics.

The module consists of three components:
(1) MLP1 (VO Velocity Estimator): a supervised network that estimates metric velocity from scaled VO poses and IMU pre-integration.
(2) State Propagation: a deterministic module that propagates the last fused state forward using IMU pre-integration and estimated velocity.
(3) MLP2 (Adaptive Fusion Policy): an RL agent that learns to weigh the IMU prediction against the VO observation to maximize long-term estimation consistency.

\textbf{Policy Formulation (for MLP2)}:
We model the fusion process as a Markov Decision Process, where each fusion step influences future drift and uncertainty. 
The state stacks VO and IMU pose/velocity estimates together with confidence cues, where VO confidence is derived from parallax and backend residuals and IMU confidence from the bias estimator and the integration horizon since the last VO update.
The policy outputs per-axis fusion weights
\[
\mathbf{w}=[w_{p_x},w_{p_y},w_{p_z},\,w_{v_x},w_{v_y},w_{v_z},\,w_q]\in[0,1]^7,
\]
Each component is interpreted as the VO mixing weight and we use a convex per-axis blend: the IMU weight is $1-w$ for the same axis.
Let $\mathbf{W}_p=\mathrm{diag}(w_{p_x},w_{p_y},w_{p_z})$ and $\mathbf{W}_v=\mathrm{diag}(w_{v_x},w_{v_y},w_{v_z})$.
Then
\[
p_k=\mathbf{W}_p\,p_k^{\text{VO}} + (\mathbf{I}-\mathbf{W}_p)\,p_k^{I},\quad
v_k=\mathbf{W}_v\,v_k^{\text{VO}} + (\mathbf{I}-\mathbf{W}_v)\,v_k^{I},
\]
and orientation is updated via $\mathrm{slerp}\!\left(q_k^{I},\,q_k^{\text{VO}},\,w_q\right)$ to preserve quaternion normalization\cite{Shoemake1985AnimatingRW}.
When VO is skipped, $\mathbf{w}$ defaults to zero and the system relies purely on IMU propagation.
The reward encourages long-term consistency rather than per-step accuracy, formulated as:
\begin{equation}
\label{eq:reward_fusion}
r_k = -\|\mathbf{p}_k - \mathbf{p}_{gt}\|_2^2
      - \lambda \, \mathrm{Tr}(\boldsymbol{\Sigma}_k)
\end{equation}

The first term drives the agent to reduce trajectory error, while the trace penalty discourages overconfident yet inaccurate updates by penalizing large predicted uncertainty.
$\boldsymbol{\Sigma}_k$ denotes a low-cost approximation of the fused-state uncertainty, 
constructed from the IMU residual standard deviation $\sigma^{\mathrm{imu}}_k$ and VO confidence $c^{\mathrm{vo}}_k$ as
$\boldsymbol{\Sigma}_k = \alpha \mathrm{diag}((\sigma^{\mathrm{imu}}_k)^2) +\mathrm{diag}((1-c^{\mathrm{vo}}_k)^2)$.

\textbf{Training Protocol}:
We adopt a two-stage training scheme: MLP1 is trained under supervision, and MLP2 is initialized with a supervised fusion loss before PPO fine-tuning. 
The PPO fine-tuning is performed on real-world trajectories (EuRoC, TUM-VI) in a Gym-style replay environment, 
where recorded IMU–VO pairs are replayed step-by-step to compute fusion rewards. 
This setup inherently captures real sensor drift and noise, eliminating the need for synthetic domain randomization.

\begin{table*}[t]
  \caption{Comparison with traditional CPU-based monocular visual-inertial odometry systems on the EuRoC MAV dataset. We report the SE(3)-aligned RMSE ATE (m). The Scale Error (\%) row for our method reports the percentage error of our initial scale estimation.}
  \label{tab:com-classic}
  \centering
  
  \begin{tabular}{@{}lcccccccccccc@{}}
    \toprule
    Method  & MH2 & MH3 & MH4 & MH5 & V11 & V12 & V13 & V21 & V22 & V23 & Avg \\
    \midrule
    
    MSCKF\cite{2013High}  & 0.45 & 0.23 & 0.37 & 0.48 & 0.34 & 0.20 & 0.67 & 0.10 & 0.16 & 1.13 & 0.413 \\
    OKVIS\cite{Leutenegger2013KeyframeBasedVS}  & 0.37 & 0.25 & 0.27 & 0.39 & 0.094 & 0.14 & 0.21 & 0.090 & 0.17 & 0.23 & 0.221 \\
    ROVIO\cite{2015Robust}  & 0.25 & 0.25 & 0.49 & 0.52 & 0.10 & 0.10 & 0.14 & 0.12 & 0.14 & 0.14 & 0.225 \\
    VINS-MONO\cite{qin2018vins}  & 0.15 & 0.22 & 0.32 & 0.30 & 0.079 & 0.11 & 0.18 & 0.080 & 0.16 & 0.27 & 0.187 \\
   
    VI-DSO\cite{von2018direct}  & 0.044 & 0.117 & 0.132 & 0.121 & 0.059 & 0.067 & 0.096 & 0.040 & 0.062 & 0.174 & 0.091 \\

    DM-VIO\cite{von2022dm}  & 0.044 & 0.097 & 0.102 & 0.096 & 0.048 & 0.045 & 0.069 & \textbf{0.029} & 0.050 & 0.114 & 0.069 \\

    ORB-SLAM3\cite{campos2021orb}  & \textbf{0.037} & \textbf{0.046} & \textbf{0.075} & \textbf{0.057} & 0.049 & \textbf{0.015} & \textbf{0.037} & 0.042 & \textbf{0.021} & \textbf{0.027} & \textbf{0.041} \\
    
    \midrule 
    
    \textbf{Ours}  & 0.064 & 0.119 & 0.112 & 0.112 & \textbf{0.047} & 0.125 & 0.073 & 0.055 & 0.036 & 0.179 & 0.092 \\
   Scale Error (\%)  & 1.1 & 1.5 & 1.4 & 1.4 & 0.1 & 1.1 & 1.5 & 0.8 & 0.5 & 1.7 & 1.11 \\
    \bottomrule
  \end{tabular}
\end{table*}

\begin{table}[t]
  \caption{Comparison with traditional CPU-based monocular visual–inertial odometry systems on the TUM-VI dataset. 
  We report SE(3)-aligned RMSE ATE (m). 
  For each scene type, the value is the average over the three corresponding sequences (1–3); full per-sequence results are provided in the Appendix D.}
  \label{tab:com-classic-tum}
  \centering
  
  \begin{tabular}{@{}lccccc}
    \toprule
    Seq & VINS & OKVIS & DM-VIO  & Ours  \\
    \midrule
    
    Corr(1-3) & 1.05 & 0.46 &  \textbf{0.30} & 0.38  \\
    Mag(1-3) & \textbf{1.90} & 2.48 &  2.09 & 1.96  \\
    Room(1-3) & \textbf{0.083} & 0.24 &  \textbf{0.083} & 0.10  \\
    Slide(1-3) & 0.74 & 1.86 &  \textbf{0.59} & 0.74  \\
    Avg&0.94&1.22&\textbf{0.77}&0.80\\

    \bottomrule
  \end{tabular}
\end{table}
\section{Experiment}

\subsection{Experimental Setup}
\label{sec:setup}
\noindent\textbf{Datasets.} 
Our visual net is first pre-trained on the large-scale synthetic TartanAir dataset~\cite{wang2020tartanairdatasetpushlimits}.
To adapt to real-world sensor characteristics, we fine-tune the IMU Bias Estimator and both RL agents 
on a single validation sequence from each dataset (MH\_01 in EuRoC\cite{article1}, Corridor\_4 in TUM-VI\cite{schubert2018vidataset}) 
and evaluate on all remaining sequences without further tuning. 
All datasets used for training and evaluation are publicly available.

\noindent\textbf{Evaluate Metrics.} We evaluate our method against baselines using three key criteria:
\begin{itemize}
    \item \textbf{Accuracy:} We report the \textbf{Absolute Trajectory Error (ATE)} in meters. This is computed after aligning the estimated trajectory with the ground truth using an $SE(3)$ similarity transformation. We also report the final \textbf{Scale Error}  to evaluate the accuracy of our initialization.
    \item \textbf{Efficiency:} We report the average \textbf{Throughput (FPS)}, this metric is computed as the total number of image frames divided by the total runtime (in seconds). 
    \item \textbf{Resources:} For our method and other deep learning-based baselines, we report the peak \textbf{GPU Memory (VRAM)} usage in Gigabytes. This is a critical metric for evaluating the feasibility of deployment on resource-constrained edge devices.
\end{itemize}

\noindent\textbf{Baselines.}
We evaluate our method against two distinct categories of state-of-the-art baselines, each serving a specific evaluation purpose: \begin{itemize} 
\item \textbf{Classical VIO (CPU-based):} We select high-performance, CPU-based methods such as VINS-Mono and ORB-SLAM3. Our comparison is intended to benchmark our final accuracy and honestly evaluate the trade-offs made to achieve high efficiency.  \item \textbf{GPU-Based VIO (Learning-based):} We compare against leading deep learning-based methods, such as iSLAM and DPVO. These serve as our direct peers, as they operate on the same GPU hardware. This comparison is  benchmark for evaluating overall performance, including accuracy, throughput, and resource consumption. \end{itemize} All baseline results are obtained by running their publicly available, open-source implementations on our evaluation hardware to ensure a consistent and fair comparison.

\noindent\textbf{Hardware.}
All experiments are conducted on a uniform platform. GPU-based methods are benchmarked on an NVIDIA RTX 3090 GPU. CPU-based methods  are benchmarked on dual Intel Xeon Platinum 8260 processors.
Analysis of deployment feasibility in Appendix C.

    
    

\begin{table}
  \caption{CPU-side breakdown per keyframe on EuRoC. 
  BA/VIBA runs on CPU for all methods. }
  \label{tab:2}
  \centering
  \small
  \begin{tabular}{@{}lccc@{}}
    \toprule
    & ORB-SLAM3  & DM-VIO  & Ours  \\
    \midrule
    Tracking & 23.22 (CPU) & 10.34 (CPU) & \textbf{9.83 (GPU)} \\
    \textbf{BA/VIBA (CPU)} & \textbf{121.09} & \textbf{26.49} & \textbf{12.77} \\
    Mapping excl. BA  & 70.41(CPU) & \textbf{27.18(CPU)} & 32.15(GPU) \\
    \midrule
    Total (ms) & 214.72 & 64.01 & \textbf{54.75} \\
    \%BA of total & 56\% & 41\% & \textbf{23\%} \\
    \bottomrule
  \end{tabular}
\end{table}
\subsection{Accuracy and Computational Analysis}
\label{sec:accuracy_analysis}

We first evaluate the final trajectory accuracy of our framework to demonstrate that our decoupled architecture can compete with state-of-the-art (SOTA) classical VIO methods. In Table~\ref{tab:com-classic} and \ref{tab:com-classic-tum}, we compare the Absolute Trajectory Error (ATE) [m] of our method against the baselines on the EuRoC and TUM-VI dataset.

The results clearly indicate that our framework's accuracy significantly surpasses filter-based approaches. Critically, when compared against optimization-based SOTA, our method  makes a clear trade-off: it does not match the gold-standard precision of systems like ORB-SLAM3, but it achieves an accuracy level comparable to other established optimization methods like DM-VIO. This demonstrates that our decoupled architecture, which mitigates the monolithic VIBA, can achieve a robust and effective level of accuracy, positioning it as a strong candidate for high-speed applications.

To examine the bottleneck fairly, we separate CPU and GPU timings. For all methods, BA/VIBA is measured on CPU only under the same CPU setup. Our approach uses a GPU front-end, which is reported separately.
Table \ref{tab:2} reports the CPU-side breakdown: compared to ORB-SLAM3 and DM-VIO, our method lowers the CPU BA/VIBA time from 121.09→12.77 ms and 26.49→12.77 ms. This indicates a structural reduction in optimization load. The total time includes GPU front-end where applicable; a device ledger is provided to keep the accounting transparent.

\begin{table*}[t]
  \caption{Efficiency and resource comparison with SOTA GPU-based VO/VIO methods on the EuRoC MAV dataset. We report SE(3)-aligned RMSE ATE (m) for VIO, Sim(3) for VO, average throughput (FPS), and peak GPU VRAM usage (GB).}
  \label{tab:3}
  \centering

  \begin{tabular}{@{}lccccccccccccc@{}}
    \toprule
    Method &   MH2 & MH3 & MH4 & MH5 & V11 & V12 & V13 & V21 & V22 &  V23 & Avg& FPS & VRAM  \\
    \small
    
    DPVO\cite{teed2023deep} &   \textbf{0.055}& 0.158& 0.137& 0.114& 0.050& 0.140& 0.086& 0.057& 0.049& 0.211 &0.106 & 22 & 4.92  \\
    iSLAM\cite{fu2024islam}  &   0.460& 0.363& 0.936& 0.478& 0.355& 0.391& 0.301& 0.452& 0.416& 1.133& 0.529 & 31 & 6.47   \\
    DROID\cite{teed2021droid}  & 0.121& 0.242& 0.399& 0.270& 0.103& 0.165& 0.158& 0.102& 0.115& 0.204&0.188 & 14 & 8.63   \\
    \textbf{Ours}   & 0.064 & \textbf{0.119} & \textbf{0.112} & \textbf{0.112} & \textbf{0.047} & \textbf{0.125} & \textbf{0.073} & \textbf{0.055} & \textbf{0.036} & \textbf{0.179}& \textbf{0.092} & \textbf{39} & \textbf{4.37}   \\
    \bottomrule
  \end{tabular}
\end{table*}
\subsection{Efficiency and Resource Comparison}
\label{sec:efficiency_analysis}

To conduct a fair evaluation of our system's efficiency, we now benchmark our framework against leading GPU-based VIO/VO methods. 
Several recent methods~\cite{yang2022efficient,9583805,Vodisch_2023_CVPR} are trained and evaluated primarily on large-scale driving benchmarks with ground-vehicle motion priors, and their released models are tailored to those settings. 
In our preliminary test, directly applying the official VS-VIO model~\cite{yang2022efficient} to EuRoC caused much larger ATE than both classical VIO and our method (see Appendix~D), highlighting the severe domain gap. We treat these driving-focused methods as conceptually related but do not include them as main numerical baselines on EuRoC/TUM-VI.

Among the GPU-based methods we re-evaluate on EuRoC, our approach achieves the best average accuracy, the highest throughput, and the lowest VRAM usage Table~\ref{tab:3}. Our framework  delivers a significantly higher average throughput of 39 FPS, which is 1.77 times faster than DPVO. This comprehensive improvement that achieves higher accuracy and higher speed simultaneously is a direct consequence of our Select Agent. Unlike methods that process every incoming frame or use a fixed skip-rate, our RL policy learns to avoid redundant computation. This drastically improves the average frame processing time without sacrificing trajectory accuracy. Furthermore, this efficiency extends to resource consumption. Our method consumes only 4.37GB of VRAM, representing a 45.2\% reduction compared to the memory-intensive DROID-VO.

These results indicate that our RL-based architecture offers a favorable accuracy–throughput–memory trade-off, making it a promising candidate for deployment on resource-constrained platforms.




\subsection{Ablation Study}
We now perform a series of ablation studies to validate the design choices of our framework.

\begin{figure}
  \centering
  
  \begin{subfigure}[b]{0.48\columnwidth}
    \includegraphics[width=\linewidth]{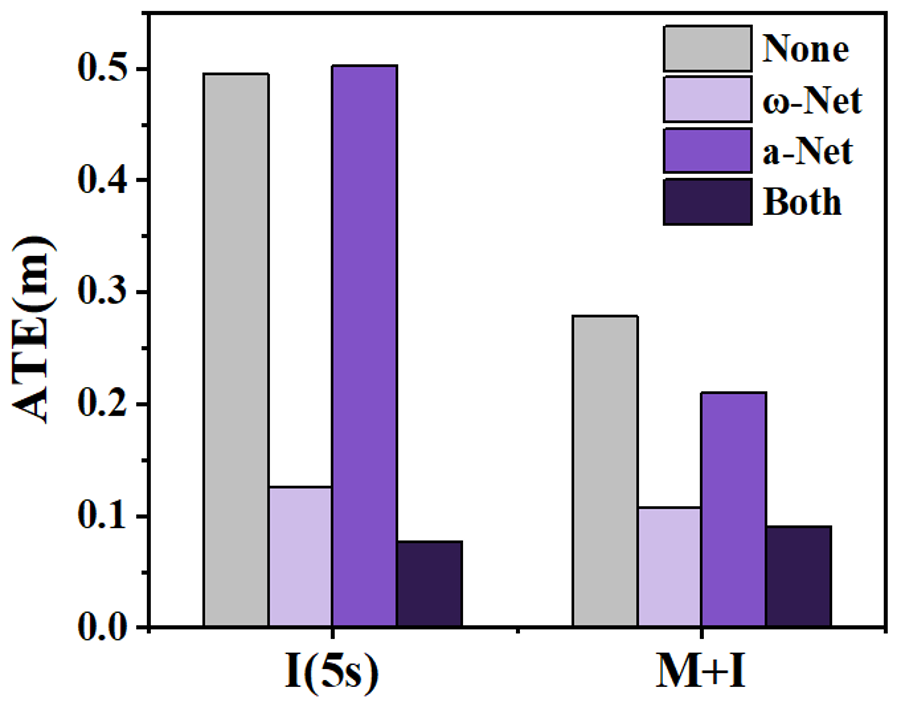}
    \caption{}
    \label{fig:ablation_bias_a}
  \end{subfigure}
  \hfill 
  \begin{subfigure}[b]{0.48\columnwidth}
    \includegraphics[width=\linewidth]{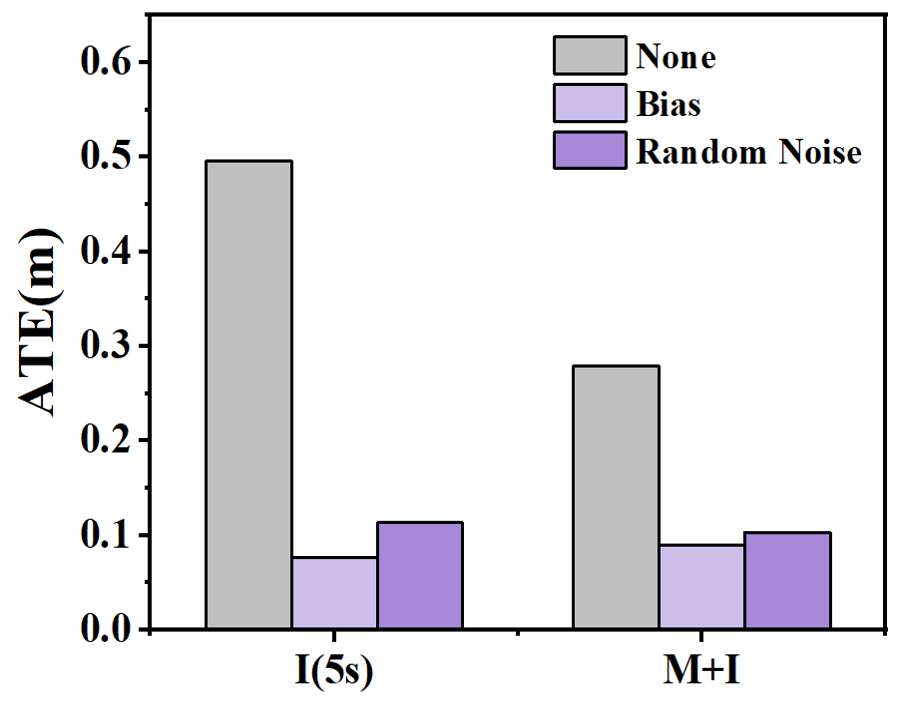}
    \caption{}
    \label{fig:ablation_bias_b}
  \end{subfigure}
  
  \caption{Ablation study for the IMU Bias Estimator. (a) Compares final ATE with different bias components enabled. (b) Compares ATE for different bias output strategies.}
  \label{fig:ablation_bias}
\end{figure}

\noindent\textbf{Validation of IMU Bias Estimator.} As our RL framework relies on high-fidelity IMU predictions , we first validate our IMU Bias Estimator (Sec.~\ref{sec:bias_estimator}). We created a test set of 100 5-second trajectory segments from EuRoC. In Figure~\ref{fig:ablation_bias_a}, we compare four estimator configurations on two metrics: pure IMU strapdown ATE (on the 5s clips) and final framework ATE. The No Bias Est baseline suffers from severe drift, while our full model (Omega + Accel) achieves the lowest ATE in both tests. This confirms that estimating both gyro and accelerometer biases is necessary for high-fidelity prediction and final system accuracy. We also validate our design choice of estimating a Fixed Bias (Sec.~\ref{sec:bias_estimator}) versus a general Random Noise  model. As shown in Figure~\ref{fig:ablation_bias_b}, the Fixed Bias model yields a lower final framework ATE. 

\begin{figure}[t]
  \centering
  
   \includegraphics[width=1\linewidth]{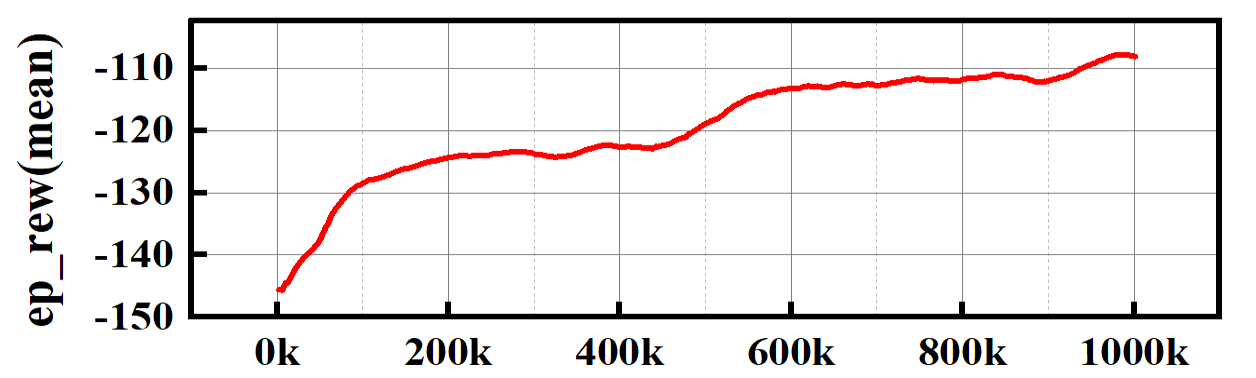}

   \caption{Training curve of the Select Agent using PPO.
   }
   \label{fig:ppo}
\end{figure}

\noindent\textbf{Ablation on RL-based Computational Scheduling.}
\label{sec:ablation_select_agent}
As shown in Fig.~\ref{fig:ppo}, the PPO-trained Select Agent converges stably, with the mean episodic reward improving consistently over 1M timesteps.
We analyze our Select Agent with two studies and introduce a strong RL baseline.

\begin{figure}
  \centering
  
  \begin{subfigure}[b]{0.48\columnwidth}
    \includegraphics[width=\linewidth]{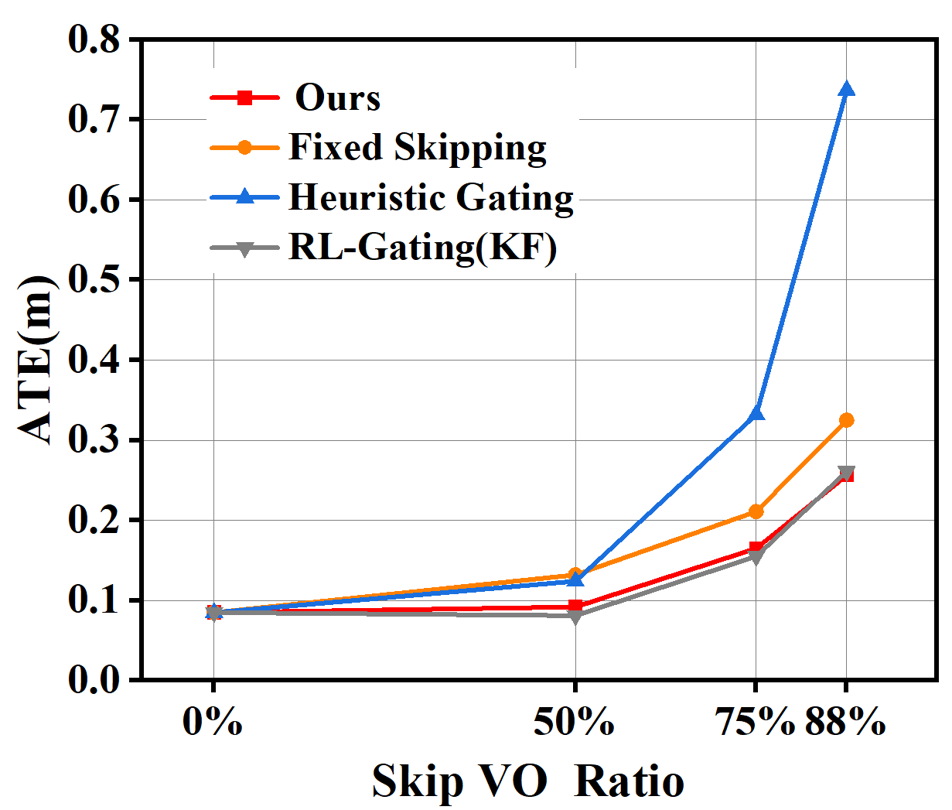}
    \caption{}
    \label{fig:ablation_select_a}
  \end{subfigure}
  \hfill 
  \begin{subfigure}[b]{0.48\columnwidth}
    \includegraphics[width=\linewidth]{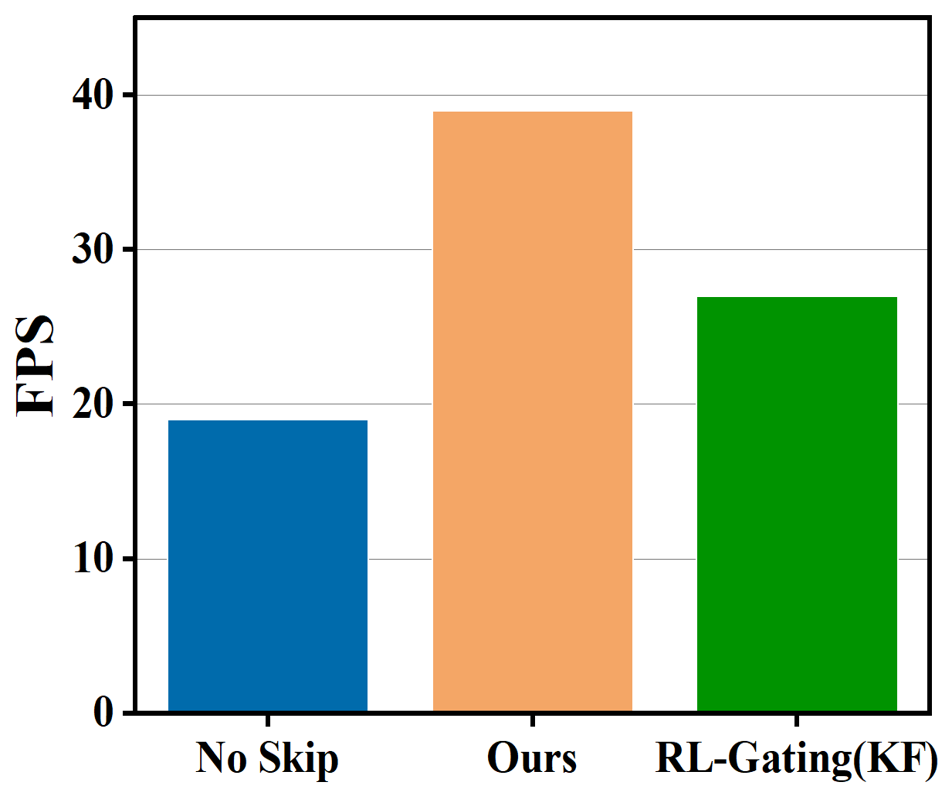}
    \caption{}
    \label{fig:ablation_select_b}
  \end{subfigure}
  
  \caption{Ablation study for the Select Agent (a) ATE vs. skip ratio comparing our IMU-only prior scheduling, fixed skipping, heuristic gating, and the RL-gating(KF) baseline (b) Throughput under a 50\% skip target: IMU-only prior scheduling attains higher FPS than RL-gating(KF) with only a marginal ATE increase.}
  \label{fig:ablation_bias}
\end{figure}

\textbf{(1) Policy vs. Fixed/Heuristic vs. RL-gating (KF).}
Beyond fixed-step skipping and heuristic gating, we implement an RL-gating(KF) baseline that uses IMU+visual inputs (I+V) to decide keyframe updates, mirroring keyframe-scheduling ideas in prior work. 
For fairness, all schedulers operate at similar average skip ratios ($\sim$50\%, $\sim$75\%, $\sim$87.5\%) by adjusting the reward coefficients in Eq.~\ref{eq:reward_sel}, and are trained under the same budget.
As shown in Fig.~\ref{fig:ablation_select_a}, accuracy at 50\% skip is comparable across learned policies, but with aggressive skipping (75–87.5\%) our IMU-only prior scheduling maintains a flatter degradation curve than heuristic gating and remains competitive with RL-gating(KF). 
This indicates the learned policy captures motion-dependent uncertainty and selectively activates VO without relying on visual features at decision time.

\textbf{(2) IMU-only vs.RL-gating (KF)\cite{messikommer2024reinforcement} efficiency.}
Fig.~\ref{fig:ablation_select_b} compares throughput under a 50\% average skip target. 
Our IMU-only policy yields substantially higher FPS because it avoids feature extraction during scheduling, while its ATE increase remains marginal ($<\!3\%$ in our setting). 
Thus, adding visual inputs provides limited incremental information for long-horizon scheduling yet incurs nontrivial compute, whereas IMU-only prior decisions preserve accuracy–latency trade-offs more efficiently.


\noindent\textbf{Ablation on RL-based Adaptive Fusion.}
\label{sec:ablation_fusion_agent}
On MH\_04, we ablate the Fusion Agent by comparing three strategies:
(i) a Heuristic Fusion baseline with a fixed 0.9 weight on VO,
(ii) an EKF Fusion baseline, and
(iii) our RL-trained agent. 
As shown in Table~\ref{tab:ablation_fusion_strat}, the RL agent consistently yields the lowest ATE, suggesting that PPO fine-tuning by directly optimizing the long-horizon drift reward in Eq.~\ref{eq:reward_fusion} learns context-dependent fusion weights that adapt to motion and VO confidence.

We further reuse the same Fusion Agent with a different VO frontend, DROID-VO, and again observe a substantial ATE reduction (Table~\ref{tab:ablation_fusion_strat}). 
This indicates that the learned policy generalizes across VO backends, as its state–action definition depends only on physical quantities rather than architecture-specific features.
\begin{table}
  \centering
  \small 
  \caption{Ablation study on the Adaptive Fusion Agent.}
  \label{tab:ablation_fusion_strat}
  
  \begin{tabular}{@{}lcc@{}}
    \toprule
    Experiment & Method & ATE (m) $\downarrow$ \\
    \midrule
    
    \multirow{3}{*}{\textbf{Strategy Comparison}}
      & Heuristic Fusion & 0.143 \\
      & EKF Fusion & 0.127 \\
      & \textbf{Ours (RL Agent)} & \textbf{0.112} \\
      
    \midrule
    \multirow{2}{*}{\textbf{Generalization}}
      & DROID-VO (Baseline) & 0.399 \\
      & DROID-VO + \textbf{Ours} & \textbf{0.237} \\
      
    \bottomrule
  \end{tabular}
\end{table}



\begin{table}[t]
  \caption{Robustness to visual degradations on EuRoC MH\_04. 5\% / 10\% of images are replaced by blurred, noise-corrupted versions.}
  \label{tab:robust_blur}
  \centering
  \begin{tabular}{@{}lccc@{}}
    \toprule
    Method & Clean & +5\% blur & +10\% blur \\
    \midrule
    DPVO~\cite{teed2023deep} & 0.137 & 0.174 & 0.192 \\
    \textbf{Ours}           & \textbf{0.112} & \textbf{0.138} & \textbf{0.153} \\
    \bottomrule
  \end{tabular}
\end{table}
\subsection{Robustness to Visual Degradations}

To assess robustness under visual quality deterioration, we synthetically corrupt the EuRoC MH\_04 sequence by randomly replacing 5\% and 10\% of images with blurred versions with added Gaussian noise, while keeping the IMU stream unchanged. 
The same corruption pattern is used for both methods. 
As reported in Table~\ref{tab:robust_blur}, the ATE of DPVO increases from 0.137\,m on the clean sequence to 0.174\,m and 0.192\,m under 5\% and 10\% blur, respectively. 
Our method also experiences some degradation, but consistently maintains lower absolute error, indicating that the learned Select and Fusion agents help the system remain accurate even when a fraction of visual observations becomes unreliable.

\section{Conclusion}




We presented a dual-agent RL-based VIO framework that reduces reliance on VIBA through learned VO scheduling and adaptive visual–inertial fusion. 
Evaluations on EuRoC and TUM-VI show that our method maintains competitive accuracy while achieving higher throughput ($1.77\times$ vs.\ DPVO) and a $9.5\times$ reduction in average backend cost compared to ORB-SLAM3. 
Future work will further weaken the dependence on optimization backends and study robustness under extrinsic miscalibration and severe VO outages.

{
  \small
  \bibliographystyle{ieeenat_fullname}
  \bibliography{main}
}

\clearpage
\setcounter{page}{1}
\maketitlesupplementary
\section*{A. Training Details}
\label{sec:appendix_training}

\subsection*{A.1. Bias Encoder Training}
Our training process is staged, following the kinematic dependencies.

\noindent\textbf{Stage 1: Gyro Bias Training.}
We first train $f_{bias}^g$ to correct the raw angular velocity:
    $\hat{\omega}(k) = \tilde{\omega}(k) - f_{bias}^g(\tilde{\Omega}, \mathbf{n}_g)$.
The corrected $\hat{\omega}(\tau)$ is integrated to predict orientation $\hat{\mathbf{q}}_t$, starting from $\hat{\mathbf{q}}_0$ ($\otimes$ is quaternion multiplication):
\begin{equation}
    \hat{\mathbf{q}}_t = \hat{\mathbf{q}}_0 + \int_{0}^{t} \frac{1}{2} \hat{\mathbf{q}}(\tau) \otimes \begin{bmatrix} 0 \\ \hat{\omega}(\tau) \end{bmatrix} d\tau
\end{equation}
Supervision is applied via the orientation error: $\mathcal{L}_{gyro} = \| \hat{\mathbf{q}}_t \ominus \mathbf{q}^{GT}_t \|^2$ ($\ominus$ denotes quaternion error)

\noindent\textbf{Stage 2: Accel Bias Training.}
 After $\mathcal{L}_{gyro}$ converges, we freeze $f_{bias}^g$ and train $f_{bias}^a$ to correct acceleration:
$\hat{a}(k) = \tilde{a}(k) - f_{bias}^a(\tilde{\mathbf{A}}, \mathbf{n}_a)$.
Using the accurate rotation $\hat{\mathbf{R}}(\tau)$ from Stage 1, we integrate $\hat{a}(\tau)$ in the world frame to predict velocity $\hat{\mathbf{v}}_t$:
\begin{equation}
    \hat{\mathbf{v}}_t = \hat{\mathbf{v}}_0 + \int_{0}^{t} \left( \hat{\mathbf{R}}(\tau) \hat{a}(\tau) - \mathbf{g} \right) d\tau
\end{equation}
Supervision is applied via the velocity L2 loss: $\mathcal{L}_{accel} = \| \hat{\mathbf{v}}_t - \mathbf{v}^{GT}_t \|^2$.
All reinforcement learning experiments were implemented using a custom Gym-style environment that simulates both the Select and Fusion Agents’ decision processes within a differentiable VIO pipeline.  
The PPO algorithm from the stable-baselines3 library was used for both agents, with separate replay buffers and reward normalization.  
Each agent interacts with the environment in an episodic fashion until convergence.

\subsection*{A.2. PPO Hyperparameter Summary}
The main hyperparameters used for PPO training are summarized in Table~\ref{tab:ppo_hyperparams}.  
Values were selected empirically for stable convergence across both agents.

The training stability of our dual-agent system is ensured through specific strategies tailored to each agent's reward structure:

\begin{itemize}
    \item \textbf{Select Agent:} Due to sparse rewards and high penalties for incorrect skipping, the training process is naturally sensitive. We stabilized convergence by employing a higher entropy coefficient to encourage exploration and a conservative learning rate, preventing premature convergence to sub-optimal policies.
    \item \textbf{Fusion Agent:} We utilize a Curriculum Learning approach. Before RL training, MLP1 undergoes supervised pre-training. This provides robust velocity priors, avoiding the instability often associated with end-to-end reinforcement learning from scratch.
\end{itemize}

\begin{table}[h]
  \centering
  \caption{PPO hyperparameters used for Select and Fusion Agents.}
  \label{tab:ppo_hyperparams}
  \small
  \begin{tabular}{lcc}
    \toprule
    \textbf{Parameter} & \textbf{Select Agent} & \textbf{Fusion Agent} \\
    \midrule
    Environment steps & 1M & 1M \\
    Learning rate & $3 \times 10^{-4}$ & $5 \times 10^{-4}$ \\
    Discount factor $\gamma$ & 0.99 & 0.99 \\
    GAE $\lambda$ & 0.95 & 0.95 \\
    Clip ratio & 0.2 & 0.2 \\
    Batch size & 64 & 64 \\
    Epochs per update & 10 & 10 \\
    Entropy coefficient & 0.05 & 0.02 \\
    Value loss coefficient & 0.5 & 0.5 \\
    \bottomrule
  \end{tabular}
\end{table}

\subsection*{A.3 Reward Design}
\paragraph{Select reward}
Recall that the Select Agent receives a terminal, episode-level reward Eq~ \ref{eq:reward_sel}.
 The scalars $A$, $B$, and $\epsilon$ control the trade-off between accuracy and computation.

\textbf{Choice of $\epsilon$, $A$, and $B$.}
On EuRoC, running DPVO with either no skipping or fixed-ratio skipping (50\%, 75\%, 87.5\%) yields ATE values typically in the range $[0.05,\,0.2]$\,m. We set $\epsilon = 0.05$, so that for ``reasonable'' trajectories the accuracy term
\[
R_{\text{acc}} = \frac{A}{\mathrm{ATE} + \epsilon}
\]
falls roughly between $4A$ and $10A$. For the same sequences, a full run contains about $4{,}000$ frames, while 50\%, 75\%, and 87.5\% skipping lead to approximately $N_f \!\approx\! 2{,}000$, $1{,}000$, and $500$ VO calls, respectively, so the cost term $R_{\text{cost}} = -B N_f$ lies in $[-4000B, -2000B]$ for typical policies.

\begin{figure}[t]
  \centering
  
   \includegraphics[width=1\linewidth]{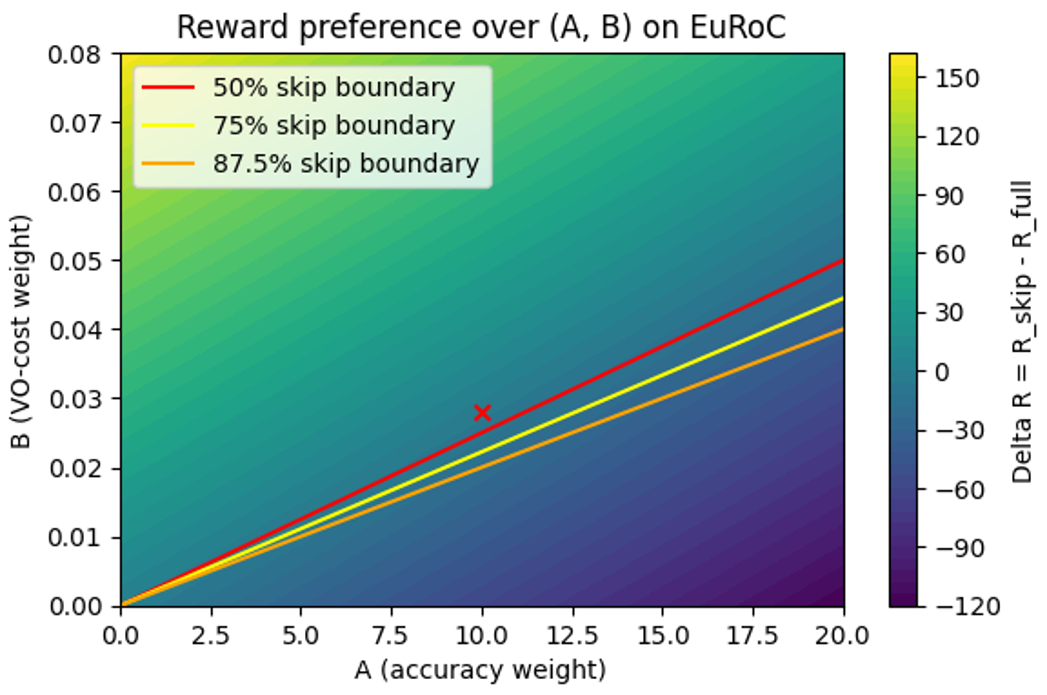}

   \caption{Fusion Agent training curve. 
   }
   \label{fig:reward}
\end{figure}

Under these simplified assumptions, the reward difference between each skipping policy and the full-frame baseline reduces to a linear function of $(A,B)$. Figure~\ref{fig:reward} visualizes these boundaries and the corresponding heatmap of $\Delta R$ over $(A,B)$, showing which regions of the parameter space favor aggressive skipping versus dense VO updates. We deliberately set $(A,B)$ slightly in favour of computation, since the dense per-step shaping term is already biased toward accuracy.We also clip the episode reward to a bounded range to avoid extreme outliers caused by occasional tracking failures.

\textbf{Dense per-step shaping.}
In addition to the terminal reward, we provide a dense shaping signal at each time step $t$ based on the instantaneous pose error:
\[
r_t^{\text{shape}} = - \mathrm{s\cdot\\RMSE}_t,
\]
where $\mathrm{RMSE}_t$ denotes the per-step position RMSE with respect to ground truth. This current-step RMSE acts as a low-variance shaping signal correlated with the final ATE, stabilizing PPO training, while the terminal reward enforces the desired long-horizon trade-off between trajectory accuracy and VO call frequency.

\paragraph{Fusion reward and uncertainty regularization.}
For the Fusion Agent, the reward has the same two-part structure: the first term directly rewards accuracy (low pose error) and is defined analogously to the Select Agent, so we omit details here. The second term penalizes the predicted uncertainty of the fused state. Concretely, we construct a diagonal covariance proxy $\Sigma_k$ from the IMU noise statistics and the VO confidence weights (as defined in the main text), and add a penalty proportional to $\mathrm{tr}(\Sigma_k)$ to the reward. This uncertainty term can be interpreted as a regularizer: it discourages the policy from relying too heavily on sensor modalities in regions where their confidence is low. In particular, when training on sequences without ground-truth trajectories, the uncertainty penalty becomes the main shaping signal, pushing the Fusion Agent toward conservative, low-uncertainty fusion patterns rather than overfitting to noisy measurements.
\begin{figure}[t]
  \centering
  
   \includegraphics[width=1\linewidth]{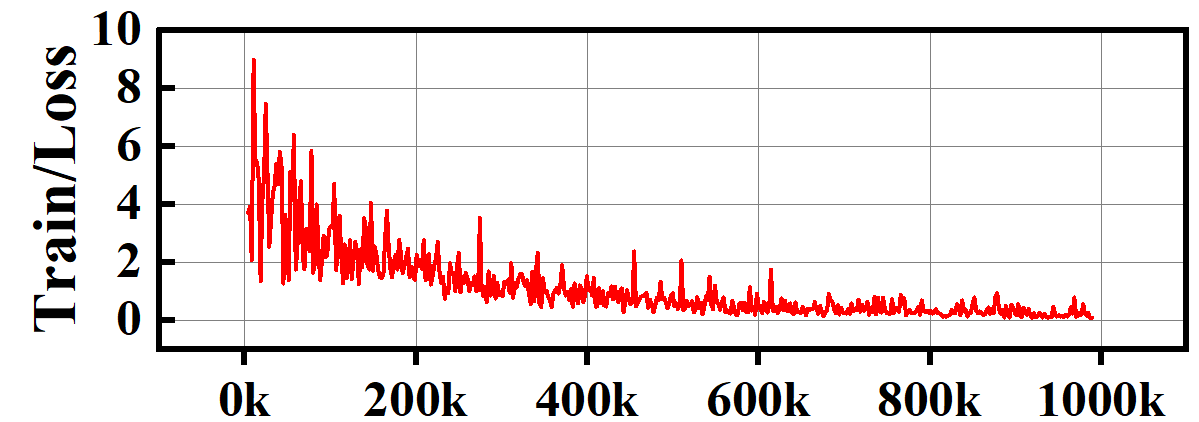}

   \caption{Reward preference over the accuracy weight A and VO-cost weight B on a EuRoC example.
   }
   \label{fig:trainfusion}
\end{figure}

\begin{figure}[t]
  \centering
  
   \includegraphics[width=1\linewidth]{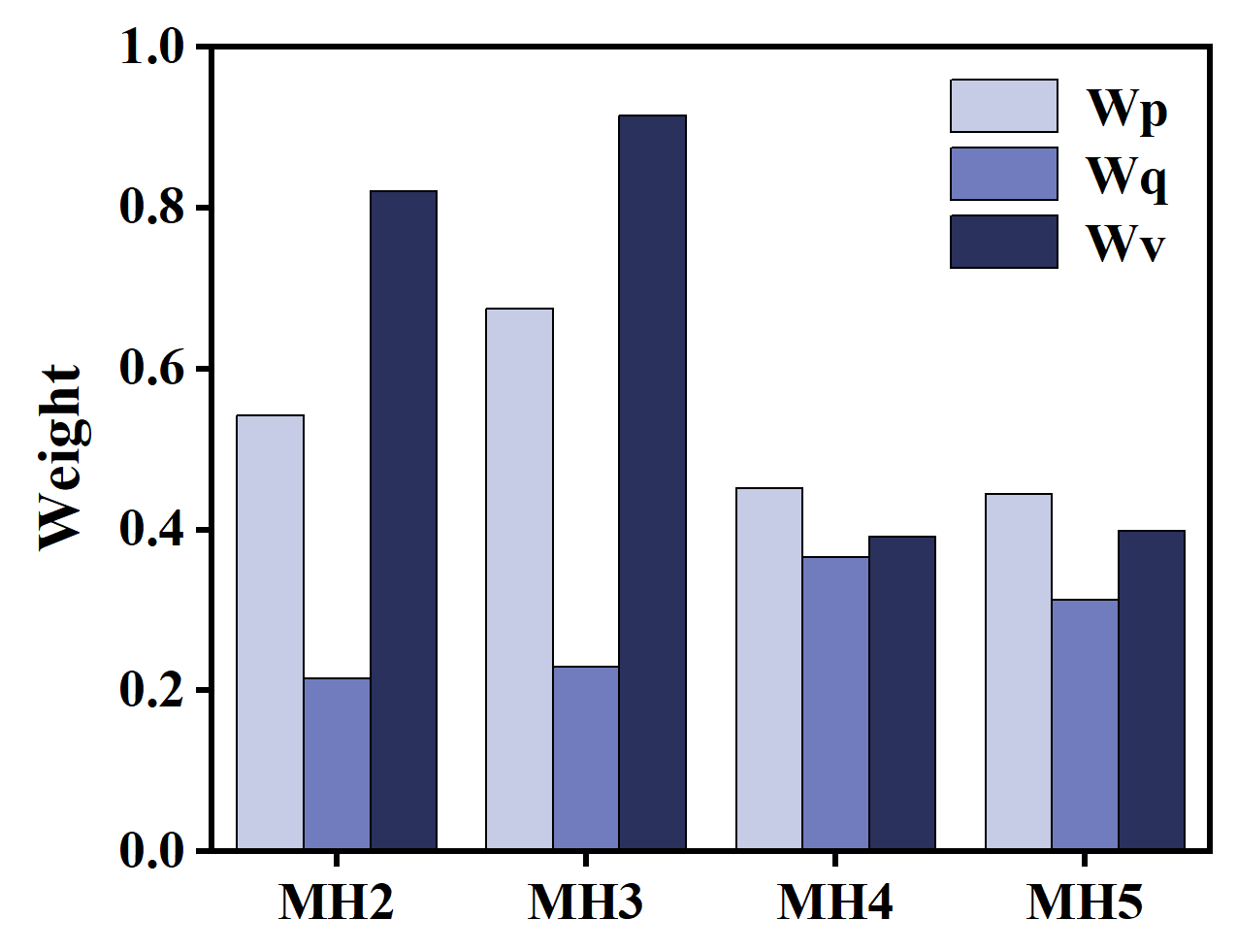}

   \caption{Visualization of learned fusion weights on EuRoC (MH2–MH5).
   }
   \label{fig:weightfusion}
\end{figure}

\subsection*{A.4 Visualization of the Fusion Agent}

Figure~\ref{fig:trainfusion} plots the training loss of the Fusion Agent over environment steps. 
After an initial warm-up phase with high variance, the loss steadily decreases and stabilizes as training progresses (up to $\sim 1$M steps), indicating that PPO converges without noticeable divergence or oscillation.

Figure~\ref{fig:weightfusion} visualizes the learned fusion weights on several EuRoC sequences (MH2--MH5). 
For each sequence, we show the average weights for position, orientation, and velocity (\(w_p, w_q, w_v\)), aggregated over all time steps. 
The agent assigns different weight patterns across sequences, reflecting how it adapts the relative contribution of each state component to the fused estimate under varying motion and visual conditions.
This behaviour illustrates that the Fusion Agent learns a context-dependent policy that adapts the IMU–visual weighting to the underlying motion and visual conditions, rather than using a fixed, hand-tuned fusion scheme.

\section*{B. IMU Pre-integration Details}

Between two consecutive camera frames $(t_k, t_{k+1})$, we receive a sequence of raw IMU readings
\[
\{(\omega^m_i, a^m_i, t_i)\}_{i=k}^{k+1}, \quad \Delta t_i = t_{i+1} - t_i,
\]
where $\omega^m_i$ and $a^m_i$ denote the measured angular velocity and linear acceleration, respectively. 
At inference time, the pre-trained bias networks $f_{\text{bias}}^g$ and $f_{\text{bias}}^a$ output gyro and accelerometer bias estimates
\[
\hat{\mathbf{b}}^g_i = f_{\text{bias}}^g(\cdot), 
\qquad
\hat{\mathbf{b}}^a_i = f_{\text{bias}}^a(\cdot),
\]
which are used to correct the raw measurements:
\begin{equation}
    \hat{\omega}_i = \omega^m_i - \hat{\mathbf{b}}^g_i, 
    \qquad 
    \hat{a}_i = a^m_i - \hat{\mathbf{b}}^a_i.
\end{equation}

We follow a standard discrete-time IMU pre-integration scheme to obtain the relative motion $(\Delta\mathbf{p}_{k}, \Delta\mathbf{q}_{k}, \Delta\mathbf{v}_{k})$ over the interval $[t_k, t_{k+1}]$. 
Let $\Delta\mathbf{R}_{k,i} \in \mathrm{SO}(3)$, $\Delta\mathbf{v}_{k,i}$, and $\Delta\mathbf{p}_{k,i}$ denote the pre-integrated rotation, velocity, and position from $t_k$ to $t_i$ in the IMU frame. 
We initialize
\[
\Delta\mathbf{R}_{k,k} = \mathbf{I},\quad 
\Delta\mathbf{v}_{k,k} = \mathbf{0},\quad
\Delta\mathbf{p}_{k,k} = \mathbf{0},
\]
and iteratively propagate for $i = k, \dots, k+1$:
\begin{equation}
 \begin{aligned}
    \Delta\mathbf{R}_{k,i+1} &= \Delta\mathbf{R}_{k,i}\,\Exp\!\big(\hat{\omega}_i\,\Delta t_i\big),\\
    \Delta\mathbf{v}_{k,i+1} &= \Delta\mathbf{v}_{k,i} + 
        \Delta\mathbf{R}_{k,i}\,\hat{a}_i\,\Delta t_i, \\
    \Delta\mathbf{p}_{k,i+1} &= \Delta\mathbf{p}_{k,i} + 
        \Delta\mathbf{v}_{k,i}\,\Delta t_i 
        + \frac{1}{2}\,\Delta\mathbf{R}_{k,i}\,\hat{a}_i\,\Delta t_i^2, 
 \end{aligned}
\end{equation}
where $\Exp(\cdot)$ is the exponential map from $\mathfrak{so}(3)$ to $\mathrm{SO}(3)$. 
At the end of the interval, we obtain
\[
\Delta\mathbf{R}_{k} \triangleq \Delta\mathbf{R}_{k,k+1},\quad
\Delta\mathbf{v}_{k} \triangleq \Delta\mathbf{v}_{k,k+1},\quad
\Delta\mathbf{p}_{k} \triangleq \Delta\mathbf{p}_{k,k+1}.
\]
The rotation $\Delta\mathbf{R}_{k}$ is converted to a unit quaternion $\Delta\mathbf{q}_{k}$ (and to axis–angle form when constructing the RL observations).

In summary, for each interval $(t_k, t_{k+1})$ the module outputs 
$(\Delta\mathbf{p}_{k}, \Delta\mathbf{q}_{k}, \Delta\mathbf{v}_{k})$ and 
bias estimates $\hat{\mathbf{b}}_k = (\hat{\mathbf{b}}^g_k, \hat{\mathbf{b}}^a_k)$,
which are fed to the Select and Fusion Agents as part of their state.

\section*{C. Analysis of Deployment Feasibility}
\label{sec:deployment_analysis}

We analyze our architecture's feasibility based on two key factors: 
First, we analyze the model size, the total combined model size for our entire VIO framework is only 32.1 MB, making them trivial to store and load on resource-constrained edge devices. Second, these new modules introduce negligible computational overhead. The inference cost of these small networks is minimal (typically $<$ 1ms), our framework does not add any new, computationally heavy layers to the main loop, and our architecture's primary efficiency gain comes from intelligent computational gating to reduces the total average FLOPs and inference time required per frame.

\section*{D. Additional Experimental Results}

\noindent\textbf{Comparison with classical CPU-based VIO on TUM-VI.}
Table~\ref{tab:com-classic-tum} reports SE(3)-aligned RMSE ATE on TUM-VI for classical CPU-based monocular VIO systems.
Our method achieves accuracy comparable to DM-VIO on average, while clearly outperforming VINS and OKVIS on most sequences.

\begin{table}[ht]
  \caption{Comparison with classical CPU-based monocular visual-inertial odometry systems on the TUM-VI dataset. We report SE(3)-aligned RMSE ATE (m).}
  \label{tab:com-classic-tum}
  \centering
  \small
  \begin{tabular}{@{}lccccc@{}}
    \toprule
    Seq & VINS & OKVIS & DM-VIO  & Ours & Length \\
    \midrule
    Corr1 & 0.63 & 0.33 &  0.19 & 0.23 & 305 \\
    Corr2 & 0.95 & 0.47 &  0.47 & 0.39 & 322 \\
    Corr3 & 1.56 & 0.57 &  0.24 & 0.52 & 300 \\
    Mag1  & 2.19 & 3.49 &  2.35 & 3.12 & 918 \\
    Mag2  & 3.11 & 2.73 &  2.24 & 2.21 & 561 \\
    Mag3  & 0.40 & 1.22 &  1.69 & 0.56 & 566 \\
    Room1 & 0.07 & 0.06 &  0.03 & 0.11 & 146 \\
    Room2 & 0.07 & 0.11 &  0.13 & 0.10 & 142 \\
    Room3 & 0.11 & 0.07 &  0.09 & 0.09 & 135 \\
    Slide1& 0.68 & 0.86 &  0.31 & 0.55 & 289 \\
    Slide2& 0.84 & 2.15 &  0.87 & 0.91 & 299 \\
    Slide3& 0.69 & 2.58 &  0.60 & 0.76 & 383 \\
    \midrule
    Avg   & 0.94 & 1.22 &  0.77 & 0.80 & 363.8 \\
    \bottomrule
  \end{tabular}
\end{table}

\noindent\textbf{Cumulative Component Analysis.}
Finally, to provide a holistic view of our contributions, Table~\ref{tab:6} reports the cumulative impact of the three main components (bias encoder, Fusion Agent, Select Agent) on EuRoC and TUM-VI.

\begin{table}[ht]
  \caption{Cumulative contribution of each component. Removing a component from the full system (ALL) degrades accuracy or efficiency.}
  \label{tab:6}
  \centering
  \small
  \begin{tabular}{@{}lccc@{}}
    \toprule
    Component      & ATE (EuRoC) & ATE (TUM-VI) & FPS \\
    \midrule
    ALL            & 0.092 & 0.80 & 39 \\
    \,- Bias Encoder & 0.279 & 1.13 & 40 \\
    \,- Fusion Agent & 0.133 & 0.94 & 39 \\
    \,- Select Agent & 0.087 & 0.76 & 21 \\
    \bottomrule
  \end{tabular}
\end{table}

Removing the bias encoder leads to a large increase in ATE on both datasets, confirming its importance for stable pre-integration.
Disabling the Fusion Agent also degrades accuracy, whereas removing the Select Agent mainly hurts efficiency (FPS drops from 39 to 21), illustrating the complementary roles of the three components.

\begin{table}[ht]
  \caption{Cross-dataset transfer performance of Visual-Selective-VIO (VS-VIO) on EuRoC MAV. We report SE(3)-aligned RMSE ATE (m) on four sequences (MH\_02–MH\_05) using the official KITTI-pretrained model without retraining.}
  \label{tab:vs}
  \centering
  \small
  \begin{tabular}{@{}lcccc@{}}
    \toprule
    Seq      & MH2 & MH3 & MH4&MH5 \\
    \midrule
    VS-VIO      & 1.134 & 1.032 & 1.029&1.355 \\
    
    \bottomrule
  \end{tabular}
\end{table}

\noindent\textbf{Additional baseline: Visual-Selective-VIO on EuRoC.}
To further assess adaptive VIO methods trained on driving data, we evaluate Visual-Selective-VIO (VS-VIO) on EuRoC without any retraining.
We reuse the official KITTI-pretrained model and convert EuRoC sequences MH\_02–MH\_05 into the KITTI-style format used by the original code, including nearest-neighbor timestamp alignment between images and ground-truth and IMU interpolation as described in Sec.~\ref{sec:setup}.
Table~\ref{tab:vs} reports the SE(3)-aligned RMSE ATE (m).
Compared to both classical VIO and our method (see Tables~\ref{tab:com-classic} and~\ref{tab:com-classic-tum}), VS-VIO exhibits substantially larger trajectory errors on these aggressive 6-DoF MAV sequences, confirming a strong domain gap between ground-vehicle training data and EuRoC.
We therefore treat VS-VIO as a conceptually related method, but do not include it as a main numerical baseline in the core results tables.

\noindent\textbf{Evaluation on failure modes.} 
We conduct stress tests on the EuRoC dataset (MH\_04) under three extreme conditions: (1) VO Outage, (2) Severe Blur, and (3) IMU Noise. The results in Table~\ref{tab:failure_analysis} demonstrate that while accuracy degrades under extreme stress, our method significantly outperforms the DPVO baseline. This resilience is primarily attributed to the \textbf{Fusion Agent}, which dynamically detects low visual reliability and suppresses visual weights, effectively leveraging the IMU Bias Estimator as a safety net.

Despite the improved robustness, residual drift in VO-denied areas suggests two directions for future work: (1) incorporating long-range temporal context into confidence estimation, and (2) establishing a feedback mechanism to re-initialize the VO backend using IMU states during prolonged outages.

\begin{table}[h]
\centering
\footnotesize
\setlength{\tabcolsep}{4pt}
\caption{Robustness Stress Test under Severe Sensor Degradation.}
\label{tab:failure_analysis}
\begin{tabular}{l|c|c|c|c}
\hline
\textbf{Method} & \textbf{Nominal} & \textbf{VO Outage} & \textbf{Severe Blur} & \textbf{IMU Noise} \\
 & (Clean Data) & (Blackout 2s) & (30\% Frames) & ($2\times$ Noise) \\
\hline
DPVO & 0.106 & 1.928 & 1.122 & - \\
\textbf{Ours} & \textbf{0.092} & \textbf{1.114} & \textbf{0.472} & \textbf{0.128} \\
\hline
\end{tabular}
\end{table}

\noindent\textbf{Scale Robustness and Monitoring.} 
Regarding scale estimation, we acknowledge that our base MLP1 primarily performs velocity smoothing and has limited authority to correct large initial scale errors. To address this, we integrated a lightweight \textbf{Online Scale Monitor}. This module calculates the motion magnitude ratio between IMU pre-integration and VO over a sliding window (15 frames). If a persistent deviation is detected, a correction factor realigns the VO scale with the IMU’s metric reference. As shown in Table~\ref{tab:scale}, under $\pm 20\%$ initialization error, our system with the Monitor effectively mitigates the scale drift.

\begin{table}[h]
    \centering
    \small
    \setlength{\tabcolsep}{6pt}
    \caption{Scale Robustness Test (ATE [m]).} 
    \label{tab:scale}
    \begin{tabular}{l|c|c|c}
        \toprule
        \textbf{Method} & \textbf{Nominal} & \textbf{Small Init} & \textbf{Large Init} \\
        \midrule
        Ours (Base) & \textbf{0.092} & 0.661 & 0.682 \\
        \textbf{Ours (+Monitor)} & 0.096 & \textbf{0.522} & \textbf{0.508} \\
        \bottomrule
    \end{tabular}
\end{table}

\noindent\textbf{Generalization:} To assess performance with minimal domain adaptation, we evaluated a model fine-tuned only on TUM-VI (Corridor\_4) applied to EuRoC (MH\_04). ATE increases (\textbf{0.092m $\to$ 0.173m}) due to the domain gap in IMU noise profiles. However, the system successfully completes the trajectory, validating robustness to unseen sensors.

\section*{E. Discussion on Algorithmic Novelty}

To further clarify our contributions relative to existing RL-based Visual Odometry (VO) works, we emphasize two fundamental distinctions in our dual-agent architecture:

\noindent\textbf{1. Paradigm Shift: ``Shift Left'' Pre-emptive Gating.} 
Traditional RL-based VO methods typically focus on post-processing, such as filtering keyframes or weighting features after they have been extracted. This represents a sunk cost where significant computation has already been spent. In contrast, our \textbf{Select Agent} proposes a ``Shift Left'' strategy: it performs pre-emptive gating on raw sensor data streams. By making skipping decisions before the visual frontend is invoked, we bypass the entire tracking and bundle adjustment pipeline for redundant frames.

\noindent\textbf{2. First RL-based VIO Fusion.} 
While prior works often use RL for visual parameter tuning, our framework is, to the best of our knowledge, the first to employ RL for dynamic IMU-Visual fusion. Our \textbf{Fusion Agent} goes beyond simple weighting; it learns a context-dependent policy that adapts the IMU-visual contribution based on motion dynamics and visual reliability.

\section*{F. Limitations and Future Work}

Although our dual-agent framework achieves a favourable accuracy--efficiency trade-off across EuRoC and TUM-VI, several aspects of the current study remain limited in scope and open up promising directions for future work.

\paragraph{Dataset and platform coverage.}
Our RL policies are trained and evaluated on EuRoC and TUM-VI using offline, log-driven environments, and thus implicitly adapt to the motion patterns, sensor characteristics, and visual statistics of these datasets. 
We do not yet explore cross-dataset transfer or deployment on substantially different platforms (e.g., wheeled robots, handheld AR devices, or cameras/IMUs with different noise profiles), which would be important to fully assess generalization.

\paragraph{Dependence on the VO backend and initialization.}
The framework assumes a reasonably calibrated VO backend and a successful initial scale estimate.
In challenging conditions with persistent VO degradation or severe miscalibration, the current policies may still fail to recover, as they are not explicitly trained to handle hard re-initialization events.
Designing agents that can detect such failure modes and proactively trigger re-initialization or fallback behaviours is an interesting extension.

\paragraph{Hardware evaluation.}
All experiments are conducted on a desktop-class CPU/GPU platform.
While we show that the proposed scheduling reduces the load on the bundle adjustment module and improves throughput, we have not yet performed a systematic study on embedded or low-power hardware, nor an explicit analysis of latency, energy, and memory footprints.
A more comprehensive hardware evaluation would be needed to fully validate the benefits for real-world onboard systems.


\paragraph{Future work.}
A natural next step is to push the framework towards real edge deployment.
On the algorithmic side, we aim to incorporate hardware constraints such as latency, power, and memory directly into the reward design, and to train and evaluate the policies on embedded platforms (e.g., Jetson-class devices) under realistic compute budgets.
On the system side, extending the scheduling mechanism to coordinate multiple perception modules (VO, mapping, semantics) sharing the same edge device, and developing even lighter-weight agent variants that can run at high frequency on low-power hardware, are promising directions.
Ultimately, our goal is to turn the proposed dual-agent controller into a drop-in, resource-aware VIO front-end for mobile robots, drones, and AR devices operating under tight real-time and power constraints.


\end{document}